\documentclass[11pt]{article}

\usepackage[final]{acl}

\makeatletter
\ifdefined\linenumbers
  \RemoveFromHook{build/column/before}[lineno]%
  \AtBeginDocument{%
    \def\@LN@makecol{%
      \@LN@orig@makecol
      \setbox\@outputbox\vbox{%
        \boxmaxdepth\@maxdepth
        \protected@write\@auxout{}{\string\def\string\@LN@column{\if@firstcolumn1\else2\fi}}%
        \box\@outputbox}}}%
\fi
\makeatother

\usepackage{times}
\usepackage{latexsym}
\usepackage[T1]{fontenc}
\usepackage[utf8]{inputenc}
\usepackage{microtype}
\usepackage{inconsolata}
\usepackage{graphicx}
\usepackage{booktabs}
\usepackage{float}
\usepackage{xspace}
\xspaceaddexceptions{]\,}
\usepackage{amsmath}
\usepackage{amssymb}
\usepackage{enumitem}
\usepackage{xcolor}
\usepackage{tikz}
\usetikzlibrary{positioning,calc,arrows.meta}
\usepackage{fontawesome5}

\definecolor{blueC}{HTML}{2A78D6}
\definecolor{orangeC}{HTML}{EB6834}
\definecolor{ink}{HTML}{0B0B0B}
\definecolor{inkgray}{HTML}{52514E}
\definecolor{grayC}{HTML}{9A9994}  
\colorlet{blue}{blueC}
\colorlet{orange}{orangeC}

\newcommand{\RQ}[1]{\textbf{RQ#1}}

\newcommand{\NTrain}{17{,}307\xspace}
\newcommand{\NSample}{1{,}000\xspace}
\newcommand{\NOpen}{18\xspace}
\newcommand{\NProp}{4\xspace}
\newcommand{\NModels}{22\xspace}
\newcommand{\WordsMedian}{345\xspace}
\newcommand{\BestModel}{GLM-5.2\xspace}
\newcommand{\BestQWK}{0.604\xspace}

\newcommand{\WorstModel}{gpt-oss-20b\xspace}
\newcommand{\WorstQWK}{0.268\xspace}
\newcommand{\MedianQWK}{0.481\xspace}
\newcommand{\GemmaQWK}{0.578\xspace}
\newcommand{\GemmaCost}{\$0.288\xspace}
\newcommand{\GemmaDiffLo}{\ensuremath{-}0.007\xspace}
\newcommand{\GemmaDiffHi}{0.061\xspace}

\newcommand{\RhoTotal}{0.37\xspace}
\newcommand{\PTotal}{0.14\xspace}
\newcommand{\NScale}{18\xspace}
\newcommand{\RhoActive}{0.58\xspace}
\newcommand{\PActive}{0.04\xspace}
\newcommand{\NActive}{13\xspace}
\newcommand{\RhoCost}{0.41\xspace}
\newcommand{\PCost}{0.21\xspace}

\newcommand{\CheapModel}{gpt-oss-20b\xspace}
\newcommand{\CheapCost}{\$0.116\xspace}

\newcommand{\CostlyModel}{Kimi-K2.6\xspace}
\newcommand{\CostlyCost}{\$7.20\xspace}
\newcommand{\CostRatio}{62\xspace}
\newcommand{\OutTokMinModel}{Gemma-4-31B\xspace}
\newcommand{\OutTokMin}{62\xspace}
\newcommand{\OutTokMaxModel}{Kimi-K2.6\xspace}
\newcommand{\OutTokMax}{1{,}862\xspace}
\newcommand{\InTokMed}{1{,}315\xspace}
\newcommand{\NTableOne}{11\xspace}
\newcommand{\NTableOmit}{7\xspace}
\newcommand{\SampleDistMaxDiff}{1.7\xspace}
\newcommand{\NTokEssays}{100\xspace}
\newcommand{\NTokModels}{11\xspace}
\newcommand{\AdjMin}{0.82\xspace}
\newcommand{\AdjMax}{0.95\xspace}
\newcommand{\ExactMin}{0.36\xspace}
\newcommand{\ExactMax}{0.49\xspace}
\newcommand{\HumanMid}{85\%\xspace}
\newcommand{\ModelMid}{97\%\xspace}
\newcommand{\SDRatio}{0.62\xspace}
\newcommand{\RecHigh}{3\%\xspace}
\newcommand{\RecLow}{8\%\xspace}
\newcommand{\NHigh}{72\xspace}
\newcommand{\NLow}{79\xspace}
\newcommand{\PooledOne}{2.15\xspace}
\newcommand{\PooledSix}{4.00\xspace}
\newcommand{\KimiBias}{\ensuremath{-}0.01\xspace}
\newcommand{\BiasMinModel}{gpt-oss-20b\xspace}
\newcommand{\BiasMin}{\ensuremath{-}0.49\xspace}
\newcommand{\BiasMaxModel}{DeepSeek-V4-Pro\xspace}
\newcommand{\BiasMax}{0.28\xspace}
\newcommand{\NBiasNeg}{12\xspace}
\newcommand{\DSFlashBias}{\ensuremath{-}0.04\xspace}
\newcommand{\DSProBias}{0.28\xspace}
\newcommand{\QwenParse}{22\%\xspace}

\newcommand{\ClaudeBest}{Claude Fable 5.1\xspace}
\newcommand{\ClaudeBestQWK}{0.748\xspace}
\newcommand{\ClaudeBestLo}{0.588\xspace}
\newcommand{\ClaudeBestHi}{0.853\xspace}
\newcommand{\SubOpenBest}{MiniMax-M2.7\xspace}
\newcommand{\SubOpenBestQWK}{0.610\xspace}
\newcommand{\OpusQWK}{0.734\xspace}
\newcommand{\HumanHumanQWK}{0.745\xspace}
\newcommand{\RhoLenHuman}{0.73\xspace}
\newcommand{\RhoLenPredMed}{0.42\xspace}
\newcommand{\RhoLenPredMin}{0.24\xspace}
\newcommand{\RhoLenPredMax}{0.51\xspace}
\newcommand{\RhoLenErrMed}{\ensuremath{-}0.56\xspace}
\newcommand{\NLenErrNeg}{18\xspace}
\newcommand{\ErrQOne}{0.52\xspace}

\newcommand{\ErrQFiveAbs}{0.78\xspace}
\newcommand{\MedianModelHuman}{0.481\xspace}
\newcommand{\WordsQOneHi}{234\xspace}
\newcommand{\WordsQFiveLo}{478\xspace}
\newcommand{\HumanQOne}{1.88\xspace}
\newcommand{\HumanQFive}{4.07\xspace}
\newcommand{\PromptCoverage}{90\%\xspace}
\newcommand{\PromptCoverageTrain}{90\%\xspace}
\newcommand{\PromptBest}{Car-free cities\xspace}
\newcommand{\PromptBestQWK}{0.575\xspace}
\newcommand{\PromptWorst}{Driverless cars\xspace}
\newcommand{\PromptWorstQWK}{0.382\xspace}
\newcommand{\MMMed}{0.529\xspace}
\newcommand{\MMMin}{0.209\xspace}
\newcommand{\MMMax}{0.768\xspace}
\newcommand{\EnsThree}{0.594\xspace}
\newcommand{\EnsFive}{0.584\xspace}
\newcommand{\EnsAll}{0.535\xspace}
\newcommand{\NHighParse}{17\xspace}
\newcommand{\MMWithin}{0.630\xspace}
\newcommand{\MMAcross}{0.523\xspace}
\newcommand{\RhoConv}{0.93\xspace}
\newcommand{\PConv}{0.001\xspace}
\newcommand{\RhoGap}{\ensuremath{-}0.86\xspace}
\newcommand{\PGap}{0.014\xspace}
\newcommand{\GapMin}{0.08\xspace}
\newcommand{\GapMax}{0.40\xspace}

\newcommand{\RhoLenHumanPromptMin}{0.58\xspace}
\newcommand{\RhoLenHumanPromptMax}{0.80\xspace}

\newcommand{\CalSingleHundred}{0.538\xspace}
\newcommand{\CalTopThreeHundred}{0.621\xspace}
\newcommand{\CalLenHundred}{0.688\xspace}
\newcommand{\CalCombHundred}{0.758\xspace}
\newcommand{\CalLenTwentyFive}{0.648\xspace}
\newcommand{\CalCombTwentyFive}{0.716\xspace}
\newcommand{\CalCombFourHundred}{0.771\xspace}
\newcommand{\CalCombFourHundredLo}{0.747\xspace}
\newcommand{\CalCombFourHundredHi}{0.798\xspace}
\newcommand{\CalLenFourHundred}{0.706\xspace}

\newcommand{\CalSingleFourHundred}{0.601\xspace}
\newcommand{\CalReps}{40\xspace}
\newcommand{\CalCombRecHigh}{71\%\xspace}
\newcommand{\CalCombRecLow}{45\%\xspace}
\newcommand{\CalRawRecHigh}{9\%\xspace}
\newcommand{\CalRawRecLow}{21\%\xspace}
\newcommand{\CalCombSev}{3\%\xspace}
\newcommand{\CalRawSev}{7\%\xspace}

\newcommand{\CalCombMid}{82\%\xspace}
\newcommand{\LopoComb}{0.74\xspace}
\newcommand{\LopoCombMin}{0.63\xspace}
\newcommand{\LopoCombMax}{0.80\xspace}
\newcommand{\LopoLen}{0.67\xspace}
\newcommand{\LopoLenMin}{0.49\xspace}

\newcommand{\LopoRaw}{0.58\xspace}
\newcommand{\NLopoCombAbove}{5\xspace}
\newcommand{\RhoRawMed}{0.53\xspace}
\newcommand{\RhoPartialMed}{0.36\xspace}
\newcommand{\RhoPartialMin}{0.20\xspace}
\newcommand{\RhoPartialMax}{0.46\xspace}
\newcommand{\RhoLenGivenLLM}{0.64\xspace}
\newcommand{\SevMin}{5\%\xspace}
\newcommand{\SevMax}{18\%\xspace}
\newcommand{\SevMed}{9\%\xspace}
\newcommand{\SevBest}{6\%\xspace}
\newcommand{\NCons}{36\xspace}
\newcommand{\NConsUnder}{24\xspace}
\newcommand{\NConsOver}{12\xspace}
\newcommand{\ConsHumanHigh}{64\%\xspace}

\newcommand{\ConsUnderWordsMed}{674\xspace}
\newcommand{\ConsOverWordsMed}{236\xspace}
\newcommand{\SampleWordsMed}{340\xspace}
\newcommand{\HumanDist}{79/267/372/210/62/10\xspace}
\newcommand{\HumanHighPct}{7\%\xspace}
\newcommand{\GptSmallQWK}{0.268\xspace}
\newcommand{\GptLargeQWK}{0.317\xspace}
\newcommand{\FamNemoNano}{0.28\xspace}
\newcommand{\FamNemoSuper}{0.44\xspace}
\newcommand{\FamNemoUltra}{0.49\xspace}
\newcommand{\FamGptSmall}{0.27\xspace}
\newcommand{\FamGptLarge}{0.32\xspace}
\newcommand{\FamDSFlash}{0.47\xspace}
\newcommand{\FamDSPro}{0.47\xspace}
\newcommand{\FamGLMTwo}{0.60\xspace}
\newcommand{\FamGLMThree}{0.50\xspace}
\newcommand{\FamMMTwo}{0.50\xspace}
\newcommand{\FamMMThree}{0.39\xspace}
\newcommand{\FamKimiFive}{0.43\xspace}
\newcommand{\FamKimiSix}{0.49\xspace}
\newcommand{\SizeRatio}{24\xspace}
\newcommand{\CostRatioBestGemma}{6.4\xspace}
\newcommand{\ClassCostLo}{\$0.017\xspace}
\newcommand{\ClassCostHi}{\$1.08\xspace}
\newcommand{\SubCIHalf}{0.18\xspace}
\newcommand{\LowActiveMin}{3\xspace}
\newcommand{\LowActiveMax}{5.1\xspace}

\title{Close in Order, Off in Scale:\\
Calibrating Zero-Shot LLM Essay Scores with Human Ratings}

\author{
  Kai Rangel\,\href{https://orcid.org/0009-0008-3769-8787}{\textcolor[HTML]{A6CE39}{\faOrcid}}
  \qquad
  Eduardo Davalos\,\href{https://orcid.org/0000-0001-7190-7273}{\textcolor[HTML]{A6CE39}{\faOrcid}} \\
  Department of Computer Science, Trinity University, San Antonio, TX, USA
}

\begin{document}
\maketitle

\begin{abstract}
Large language models (LLMs) are used in increasing amounts for automated essay scoring (AES),
yet evaluations typically test few proprietary models, rarely report token usage or cost, and
summarize agreement with a single coefficient. We score \NSample essays from ASAP~2.0 with \NOpen
open-weight LLMs (21B--1.6T parameters) and \NProp proprietary models using one role-based rubric
prompt with schema-constrained output. No open-weight model reaches the quadratic weighted kappa (QWK) of 0.70 required for
operational automated scoring (best QWK \BestQWK), model size is a weak predictor of agreement,
and Gemma-4-31B is statistically tied with the best model, the 744B \BestModel, at
\CostRatioBestGemma$\times$ lower token cost. All models share two
error patterns: predicted scores cluster in the middle of the scale, and they
correlate far less with essay length than human scores do (Spearman $\rho=\RhoLenPredMed$
vs.\ \RhoLenHuman). When 100 human-scored essays are used to map scores onto the human scale
(post-hoc calibration), essay length alone (QWK \CalLenHundred) outperforms every LLM, whereas
combining the three strongest LLMs with length reaches \CalCombHundred, comparable to
human--human agreement, and largely transfers to unseen prompts. Zero-shot LLM scoring is
thus limited by calibration, not model capacity. Code is available at
\url{anonymous.4open.science/r/close-in-order_off-in-scale-48D1}.
\end{abstract}

\section{Introduction}

\begin{figure*}[t]
  \centering
  \resizebox{\textwidth}{!}{
\begin{tikzpicture}[
    font=\rmfamily\scriptsize, >={Stealth[length=4pt, width=3.4pt]},
    box/.style={draw=#1, fill=#1!8, rounded corners=2.5pt, line
      width=0.6pt, align=center,
    minimum height=0.95cm, inner xsep=5pt, text=ink},
    box/.default=grayC,
    note/.style={text=inkgray, align=center, font=\rmfamily\tiny},
    flow/.style={->, line width=0.7pt, draw=inkgray},
    stage/.style={font=\rmfamily\scriptsize\bfseries, text=inkgray,
    rotate=-45, anchor=center},
    key/.style={circle, fill=#1, draw=white, line width=0.4pt, inner
    sep=0pt, minimum size=5pt},
  ]
  \node[key=blueC] (k1) at (2.2,1.35) {};
  \node[anchor=west, text=ink] at ($(k1.east)+(0.05,0)$) {Zero-shot
  LLM scoring};
  \node[key=orangeC] (k2) at (5.3,1.35) {};
  \node[anchor=west, text=ink] at ($(k2.east)+(0.05,0)$) {Calibration
  with human scores};
  \node[key=grayC] (k3) at (9.25,1.35) {};
  \node[anchor=west, text=ink] at ($(k3.east)+(0.05,0)$) {Input};

  \node[box, minimum width=2.3cm] (in) at (0,0) {Essay\\+ holistic rubric};
  \node[box=blueC, minimum width=2.3cm] (llm) at (3.3,0) {LLM\\(zero-shot)};
  \node[note, below=1pt of llm] {22 models};
  \node[box=blueC, minimum width=2.0cm] (ms) at (6.4,0) {Model score\\1--6};

  \node[box, minimum width=2.3cm] (len) at (0,-2.05) {Essay length\\(words)};
  \node[box=orangeC, minimum width=2.3cm] (few) at (3.3,-2.05) {100
  essays with\\human scores};
  \node[box=orangeC, minimum width=2.0cm] (cal) at (6.4,-2.05)
  {Calibration\\[1pt]{\tiny\color{inkgray}Top-3 mean +
  length}\\[-1pt]{\tiny\color{inkgray}$\rightarrow$ cut-points}};

  \node[box=blueC, minimum width=3.75cm, minimum height=1.5cm] (r1) at (12.85,0)
  {Zero-shot: QWK \textbf{\BestQWK}\\[-1pt]{\tiny\color{inkgray}Best
    open-weight model}\\[-2pt]
  {\tiny\color{inkgray}below the 0.70 operational bar}};
  \node[box=orangeC, minimum width=3.75cm, minimum height=1.5cm] (r2)
  at (12.85,-2.05)
  {Calibrated: QWK
    \textbf{\CalCombHundred}\\[-1pt]{\tiny\color{inkgray}Fit on 100
    human-scored essays,}\\[-2pt]
    {\tiny\color{inkgray}tested on held-out essays}\\[-2pt]
  {\tiny\color{inkgray}comparable to human--human (\HumanHumanQWK)}};
  \draw[draw=grayC, rounded corners=3.5pt, line width=0.6pt]
  ($(r1.north west)+(-0.2,0.52)$) rectangle ($(r2.south east)+(0.2,-0.2)$);
  \node[font=\rmfamily\scriptsize\bfseries, text=inkgray, anchor=south]
  at ($(r1.north)+(0,0.1)$) {Agreement with human raters};

  \draw[flow] (in) -- (llm);
  \draw[flow] (llm) -- (ms);
  \draw[flow] (in) -- (len);
  \draw[flow] (ms) -- (cal);
  \draw[flow] (few) -- (cal);
  \draw[flow] (len.south) -- ++(0,-0.38) -| (cal.south);
  \draw[flow] (ms) -- node[above, note] {Best model's score} (r1);
  \draw[flow] (cal) -- node[above, note] {Calibrated score} (r2);

  \node[stage] at (-1.8,0) {Zero-shot};
  \node[stage] at (-1.85,-2.05) {Calibration};
\end{tikzpicture}}
  \caption{\textbf{Study design.} Top: each LLM scores every essay zero-shot, given only
  the official holistic rubric. Bottom: post-hoc calibration combines the mean score of the three
  best models with essay length and maps the result onto the 1--6 scale with cut-points learned
  from 100 essays with human scores. Right: both kinds of scores are compared with the scores
  that ASAP~2.0's human raters gave the same essays; the zero-shot QWK is for the best
  open-weight model, and the calibrated QWK is measured on the held-out essays.}
  \label{fig:pipeline}
\end{figure*}
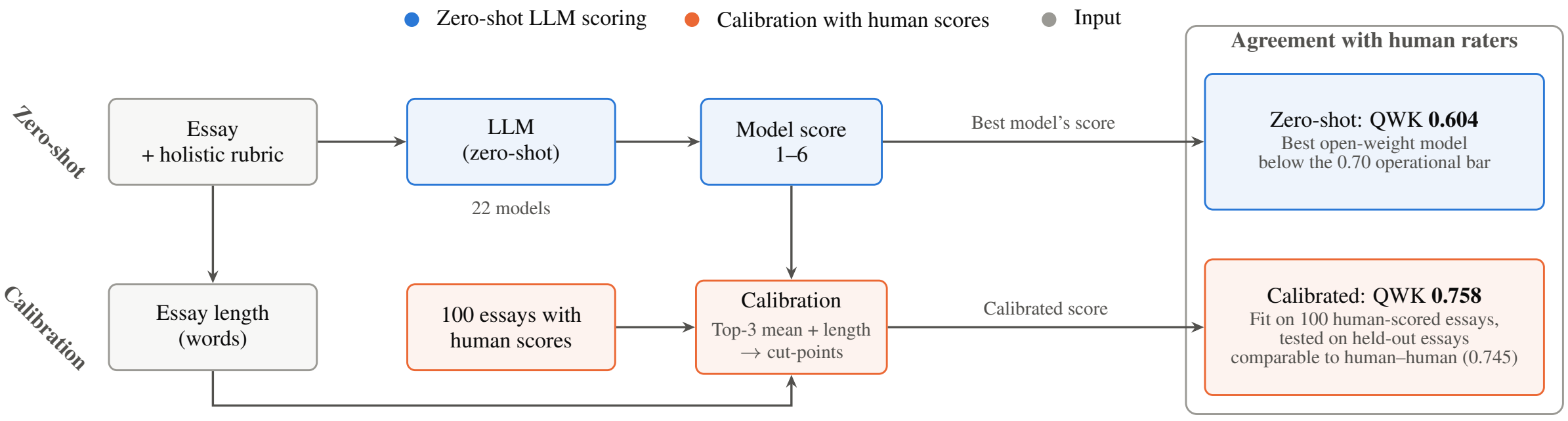

Scoring student writing is time-consuming, and large language models (LLMs) promise
immediate, rubric-based scores and feedback. The evidence for this promise is mixed. A
synthesis of 65 studies reports LLM--human agreement in quadratic weighted kappa (QWK) ranging from 0.00 to 0.97, often
below the 0.70 level required for operational automated scoring
\citep{williamson2012framework, li2025agreement},
and direct comparisons with teachers report plausible but middle-heavy and inconsistent
grades \citep{floden2025grading}. Three aspects of evaluation practice make these
results not actionable:

First, most studies evaluate a few proprietary models accessed through third-party services
\citep{li2025agreement, sessler2025grade, idowu2026specialists}. Data-governance rules,
however, often prevent schools from sending student writing to external providers, so
self-hostable open-weight models are the practically relevant class \citep{son2026sub3b}. Second, token usage and cost are rarely
reported: one multi-agent scoring
framework improves QWK but needs several model calls per essay, taking 1.7\,min per essay
compared with 0.6\,s for a single prompt \citep{jang2025roundtable}, and cost-aware evaluation
has only recently been proposed \citep{manakina2026bandit, gui2026costaware}. Third, studies
summarize agreement with a single coefficient such as QWK, which cannot show \emph{how} a
model errs. A model that shifts every score up by one point and a model that pulls all scores
toward the middle of the scale can reach similar QWK, yet only the first is corrected by
subtracting a constant. The error
pattern therefore determines whether a low-cost correction exists.

We address these gaps with a controlled comparison of \NModels LLMs on ASAP~2.0
\citep{crossley2025asap2}, holding the rubric, prompt, output schema, and essays fixed
(Figure~\ref{fig:pipeline}). We ask three research questions:
\begin{itemize}[leftmargin=2.9em,labelwidth=2.5em,labelsep=0.4em,align=left,itemsep=1pt,topsep=2pt]
  \item[\RQ{1}] How well do current open-weight LLMs agree with human raters in zero-shot
  holistic scoring, and how does agreement relate to model size and cost?
  \item[\RQ{2}] How are scoring errors distributed across score levels, essay length, and
  essay prompts?
  \item[\RQ{3}] Can ensembling or post-hoc calibration with a small set of human-scored essays
  close the gap, and how do LLMs compare with an essay-length baseline?
\end{itemize}

\noindent No open-weight model reaches QWK 0.70, and neither model size nor token cost is a
reliable predictor of agreement. All models share two error patterns: \textbf{score compression}, a concentration of predicted scores in the middle of
the scale, and \textbf{length insensitivity}, a much weaker relation between predicted
scores and essay length than in human scores. Consequently, once a few dozen
human-scored essays are used to map scores onto the human scale, essay length alone outperforms
every LLM, whereas LLM scores and length combined reach QWK \CalCombTwentyFive with 25 and
\CalCombHundred with 100 human-scored essays, comparable to human--human agreement (0.745).
Fit on other prompts, the combination reaches a mean QWK of \LopoComb on unseen essay
prompts and at least 0.70 on \NLopoCombAbove of seven. We release our code.

\section{Related Work}

Neural automated essay scoring learns prompt-specific models from human-scored essays
\citep{taghipour2016neural}; on ASAP~2.0, fine-tuned models reach QWK 0.78--0.80, above the
reported human--human agreement of 0.745 \citep{ormerod2025long}, and LLM summaries with
handcrafted features help with long essays \citep{hua2026summarization}. Zero-shot LLM scoring
needs no training data but lags behind supervised systems \citep{xiao2025human, kundu2024essay}.

Proposed improvements to LLM-based scoring include trait decomposition with min-max scaling
\citep{lee2024unleashing}, joint scoring and feedback generation \citep{stahl2024exploring},
pairwise comparison \citep{shibata2025lces}, multi-agent deliberation
\citep{jang2025roundtable, idowu2026specialists}, and structured extraction of rubric
components before scoring \citep{wang2026autoscore}. Example essays in the prompt (few-shot
examples) improve agreement substantially \citep{yancey2023rating, idowu2026specialists}, whereas more
detailed rubrics often do not \citep{yoshida2025detailed}. Light fine-tuning is also
effective: a LoRA-adapted Gemma-3-27B trained on 480 essays reaches QWK 0.70 on TOEFL11
\citep{gayed2026l1bias}. Human-in-the-loop designs keep teachers in control of the final
score \citep{xiao2025human, chu2025gradehitl}, and a recent survey reviews this design space
\citep{autograding2025survey}.

Beyond agreement, studies report systematic bias: LLMs score more harshly than human raters
\citep{kundu2024essay} or more leniently \citep{sessler2025grade}, compress grades toward
the middle \citep{floden2025grading}, and show stable negative bias on lower-order traits
\citep{kucia2026holistic}. Human scores correlate with essay length \citep{kundu2024essay},
and the signed error of small local models varies with essay length \citep{son2026sub3b}.
Work on LLM-as-a-judge finds aggregate alignment with expert judgments
\citep{chiang2023large} alongside position and verbosity biases \citep{zheng2023judging}.
Constrained decoding guarantees machine-readable output \citep{willard2023efficient},
although strict format restrictions can degrade reasoning \citep{tam2024speak}, and cost
reporting in AES is emerging, from cost--reliability curves \citep{manakina2026bandit} to
cost per defensible score \citep{gui2026costaware}.

The closest work evaluates Llama-3.1 (8B, 70B, and 405B) and gpt-oss (20B and 120B) on
ASAP~2.0 and finds holistic QWK of about 0.6 for the best model, Llama-3.1-70B
\citep{kucia2026holistic}; another benchmark evaluates 18 multimodal LLMs on trait
scoring \citep{su2025essayjudge}. We extend this line of work to \NModels current models, add
token usage and cost, and relate the error structure to a correction that requires
only a small set of human-scored essays.

\section{Methodology}

ASAP~2.0 consists of source-based argumentative essays written by U.S.\ students in grades
6--10 and scored holistically on a 1--6 scale \citep{crossley2025asap2}; it extends the
PERSUADE corpus \citep{crossley2022persuade}. From the \NTrain scored essays in the public
training split (median \WordsMedian words), we draw a reproducible random sample of \NSample
essays (seed 42) whose score distribution (\HumanDist essays for scores 1--6) mirrors the
skewed full split (Appendix Figure~\ref{fig:sampledist}). The release does not label essay prompts, so for the per-prompt analysis
we assign each essay to one of the seven prompts with keyword rules
(Appendix~\ref{app:prompts}); the rules uniquely match \PromptCoverage of the sample, and a
manual check of 21 assigned essays found no errors.

Following role-based scoring designs \citep{floden2025grading, jang2025roundtable}, every
model receives the same system role (``You are an expert essay rater\ldots{} using ONLY the
rubric provided'') and a user message containing the official holistic rubric followed by the
essay (Appendix~\ref{app:prompt}). Output is constrained by a JSON schema with an integer
\texttt{score} in $[1,6]$ and a string \texttt{rationale}, enforced by constrained decoding
where the serving stack supports it \citep{willard2023efficient}. A four-stage fallback
parser (full JSON, embedded object, \texttt{"score"} key, standalone digit) recovers
non-conforming responses; essays that remain unparseable are excluded and counted.

We evaluate \NOpen open-weight models from nine families, served through Ollama Cloud at
temperature~0: gpt-oss \citep{openai2025gptoss}, Gemma~4 \citep{gemma2026gemma4},
Nemotron~3 \citep{nvidia2025nemotron3}, DeepSeek-V4 \citep{deepseek2026v4}, MiniMax-M2.7
\citep{minimax2026m2} and M3, GLM-5.x \citep{glm2026glm5}, Kimi-K2.x \citep{kimi2025k2},
Qwen3.5, and Mistral-Large-3 (Appendix~\ref{app:models}). Model size spans 20.9B to 1.6T
total parameters. Most of these models use a \textbf{mixture-of-experts} (MoE) architecture,
which activates only a subset of its parameters for each token (e.g., 5.1B of 116.8B for
gpt-oss-120b; \citealp{openai2025gptoss}); Gemma-4-31B is a \textbf{dense} model, which uses
all of its parameters for every token \citep{gemma2026gemma4}. As a reference, we score the
first 50 essays of the same sample with \NProp proprietary models (Claude Haiku~4.5,
Sonnet~5.5, Opus~5.5, and Fable~5.1) through their vendor API at default sampling settings.

QWK is the primary metric. With $K{=}6$ score levels, observed
counts $O_{ij}$ (human score $i$, model score $j$), and expected counts $E_{ij}$ from the
outer product of the two marginal distributions,
\begin{equation}
  \kappa_w = 1 - \frac{\sum_{i,j} w_{ij} O_{ij}}{\sum_{i,j} w_{ij} E_{ij}},\qquad
  w_{ij} = \frac{(i-j)^2}{(K-1)^2}.
\end{equation}
We also report exact agreement, adjacent ($\pm$1) agreement, the share of essays with a
discrepancy of two or more points (``Disc.''), MAE, signed bias (mean predicted minus human
score), and parse rate. For cost, we rescore the first \NTokEssays essays with each
open-weight model, record the input and output tokens the API reports (output includes
reasoning tokens), and apply published per-token list prices (Appendix Table~\ref{tab:prices});
because our hosted service bills a flat fee, costs are list-price estimates. Seven models
retired from the service before this measurement are left blank. Confidence intervals (95\%) use 2{,}000 essay-level bootstrap
resamples shared across models, so differences between models are paired. Associations
across models use Spearman's $\rho$, and a partial Spearman correlation, computed after
linearly removing the rank of essay length from both variables, tests whether a model's
scores carry information beyond length.

A model may order essays correctly from weakest to strongest yet use the score scale
differently from human raters, for example by giving most essays a 3 or 4 where humans spread
them from 1 to 6. We therefore test \textbf{post-hoc calibration}: a small set of
human-scored essays is used to learn where each score level begins on a model's output, and
these boundaries are applied to the remaining essays. We repeatedly split the \NSample essays
into a calibration set of $n\in\{25,50,100,200,400\}$ human-scored essays and a held-out
test set (\CalReps random splits per $n$). Using the calibration set only, we first give every
essay a continuous score $s$: the score of the best model, the mean $\bar m$ of the three best
models (both selected by QWK), the log essay length $\ell$ (in words), or the linear
combination
\begin{equation}
  s = \beta_0 + \beta_1 \bar m + \beta_2 \ell,
\end{equation}
with $\beta$ estimated by least squares on the human scores. Five ordered
\textbf{cut-points} $c_1<\dots<c_5$ then divide $s$ into six score bands, $\hat y = 1 +
\sum_k \mathbb{1}[s > c_k]$. They start where the share of calibration essays in each band
equals the share of human scores at that level and are then moved one at a time to maximize
QWK on the calibration set. In one split with 100 calibration essays, for example, the fit is
$s = -6.91 + 0.57\,\bar m + 1.39\,\ell$ with cut-points 1.83, 2.52, 3.43, 4.08, and 5.03; a
610-word essay to which all three models assign a 3 receives $s = 3.68$, passes three
cut-points, and is scored 4 (human score: 5). The length-only variant serves as an \textbf{essay-length
baseline}. We report the mean held-out QWK with the 2.5th--97.5th percentiles across splits.
To test transfer to new essay prompts, we also fit each scoring method on the essays of six
prompts and evaluate it on the seventh (leave-one-prompt-out).

\section{Results}

\begin{table*}[t]
  \centering
  \setlength{\tabcolsep}{2.6pt}
  \begin{tabular}{@{}lrrcrrrrrr@{}}
\toprule
Model & Size (B) & QWK $\uparrow$ & 95\% CI & Exact $\uparrow$ & Disc.\ $\downarrow$ & Bias $\rightarrow\!0$ & Parsed $\uparrow$ & \shortstack[r]{Out\\tok.\ $\downarrow$} & \shortstack[r]{\$ per\\1k $\downarrow$} \\
\midrule
GLM-5.2 & 744/40$^\dagger$ & \textbf{0.604} & [0.57, 0.64] & 0.48 & 6.4\% & +0.10 & 98\% & 423 & 1.83 \\
Gemma-4-31B & 31 & 0.578 & [0.54, 0.61] & \textbf{0.49} & \textbf{5.4\%} & $-$0.09 & \textbf{100\%} & \textbf{62} & 0.29 \\
GLM-5.3-Flash & 321/18$^\ddagger$ & 0.540 & [0.50, 0.58] & 0.44 & 5.9\% & +0.22 & \textbf{100\%} & 830 & 0.47 \\
GLM-5.3 & 753/40$^\ddagger$ & 0.499 & [0.46, 0.54] & 0.45 & 6.1\% & +0.11 & \textbf{100\%} & 1{,}181 & 1.74 \\
Kimi-K2.6 & 1,040/32$^\dagger$ & 0.493 & [0.45, 0.54] & 0.45 & 6.8\% & \textbf{$-$0.01} & \textbf{100\%} & 1{,}862 & 7.20 \\
Nemotron-3-Ultra & 550$^\ddagger$ & 0.490 & [0.45, 0.53] & 0.37 & 14.2\% & $-$0.43 & \textbf{100\%} & 455 & 1.67 \\
DeepSeek-V4-Pro & 1,600/49 & 0.472 & [0.43, 0.51] & 0.41 & 8.7\% & +0.28 & \textbf{100\%} & 814 & 2.38 \\
Nemotron-3-Super & 120$^\ddagger$ & 0.441 & [0.40, 0.48] & 0.41 & 10.0\% & $-$0.13 & 99\% & 1{,}152 & 0.56 \\
MiniMax-M3 & 427$^\ddagger$ & 0.388 & [0.35, 0.43] & 0.40 & 12.5\% & $-$0.43 & \textbf{100\%} & 505 & 1.05 \\
gpt-oss-120b & 117/5.1 & 0.317 & [0.28, 0.35] & 0.40 & 14.0\% & $-$0.38 & \textbf{100\%} & 329 & 0.15 \\
gpt-oss-20b & 20.9/3.6 & 0.268 & [0.22, 0.32] & 0.36 & 17.9\% & $-$0.49 & \textbf{100\%} & 580 & \textbf{0.12} \\
\bottomrule
\end{tabular}

  \caption{\textbf{Zero-shot agreement with human raters on \NSample ASAP~2.0 essays.}
  \NTableOne of the \NOpen models, sorted by QWK; the other \NTableOmit are listed in Appendix
  Table~\ref{tab:main_omitted}, and all statistics in the text use all \NOpen. Bold: best of all
  \NOpen. Size: total/active parameters in billions; $^\dagger$taken from the family's technical report;
  $^\ddagger$not verified against a technical report. Disc.: share of essays with a
  discrepancy of two or more points. Bias: mean predicted minus human score. Parsed: share of
  essays with a valid score. Out tok.: mean output tokens per essay, including reasoning; \$ per
  1k: list-price cost per 1,000 essays (both from the first \NTokEssays essays).}
  \label{tab:main}
\end{table*}

\begin{figure*}[t]
  \centering
  \includegraphics{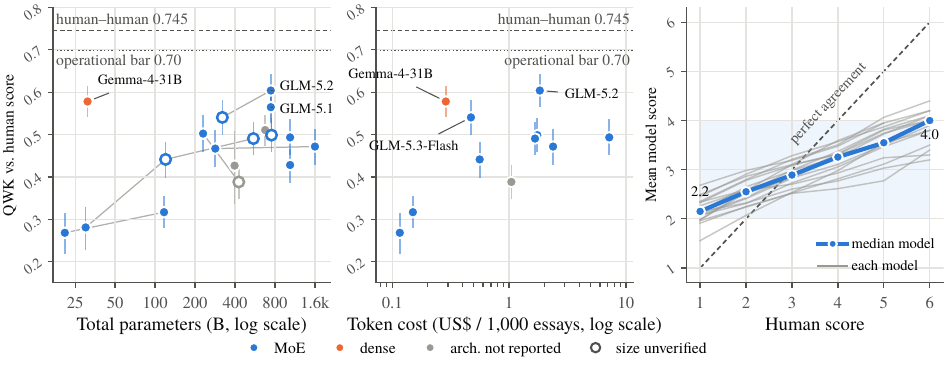}
  \caption{\textbf{Agreement versus model size and token cost, and score compression.}
  Left: QWK versus total parameters, with lines connecting models of the same family. Center:
  QWK versus list-price token cost per 1,000 essays (log scale; the \NTokModels models with
  measured usage). Both show 95\% bootstrap intervals and mark human--human agreement and
  the 0.70 operational threshold. Color gives the architecture: dense models use all parameters
  for every token, mixture-of-experts (MoE) models only a subset, and gray marks models whose
  architecture is not reported; open markers indicate a size not verified against a technical
  report. Labeled models are the three with the highest QWK in
  each panel. Right: mean predicted score at each human score level; the
  shaded band covers scores 2--4.}
  \label{fig:results}
\end{figure*}

\subsection{Agreement, Model Size, and Cost}

No open-weight model reaches the operational threshold of QWK 0.70 (\RQ{1}). QWK ranges from \WorstQWK
(\WorstModel) to \BestQWK (\BestModel), with a median of \MedianQWK (Table~\ref{tab:main}; Appendix Table~\ref{tab:main_omitted}),
and all models fall short of the human--human agreement of 0.745 reported for this dataset
\citep{ormerod2025long}. Exact agreement is only \ExactMin--\ExactMax, whereas adjacent
agreement appears high (\AdjMin--\AdjMax) because \HumanMid of essays have scores of 2--4,
where an adjacent score is easy to obtain. Our gpt-oss results (\GptLargeQWK and
\GptSmallQWK, both negatively biased) closely replicate the 0.299 and 0.261 reported
independently on ASAP~2.0 \citep{kucia2026holistic}.

Larger models are not reliably better. Across the \NScale models with known size, the rank
correlation between log total parameters and QWK is weak and not significant
($\rho=\RhoTotal$, $p=\PTotal$; Figure~\ref{fig:results}, left). The dense Gemma-4-31B scores
\GemmaQWK; a paired bootstrap of its difference from \BestModel (744B total parameters)
includes zero ([\GemmaDiffLo, \GemmaDiffHi]), and it is the only model that is not
significantly below \BestModel despite having \SizeRatio$\times$ fewer parameters. Active
parameters correlate more strongly with QWK ($\rho=\RhoActive$, $p=\PActive$, $n=\NActive$),
but this association is driven by the three lowest-scoring models, which have only
\LowActiveMin--\LowActiveMax B active parameters. Within families, larger models help only at
the lower end: Nemotron-3 rises from \FamNemoNano (Nano) to \FamNemoSuper (Super) and
\FamNemoUltra (Ultra), and gpt-oss from \FamGptSmall to \FamGptLarge, whereas DeepSeek-V4-Pro
(1.6T) scores no higher than V4-Flash (284B), \FamDSPro versus \FamDSFlash. Newer releases are
not reliably better either: GLM-5.3 (\FamGLMThree) scores below GLM-5.2 (\FamGLMTwo), and
MiniMax-M3 (\FamMMThree) below M2.7 (\FamMMTwo), whereas Kimi-K2.6 (\FamKimiSix) improves on
K2.5 (\FamKimiFive).

Token cost does not predict agreement either. Across the \NTokModels models with measured
usage, input length is similar (median \InTokMed tokens per essay),
but output length ranges from \OutTokMin tokens per essay (\OutTokMinModel) to \OutTokMax
(\OutTokMaxModel), because output tokens include the reasoning that some models generate
before the score. At list prices, scoring 1{,}000 essays costs from \CheapCost (\CheapModel)
to \CostlyCost (\CostlyModel), a \CostRatio-fold range, and the rank correlation between
cost and QWK is not significant ($\rho=\RhoCost$, $p=\PCost$; Figure~\ref{fig:results},
center). A class set of 150 essays thus costs \ClassCostLo--\ClassCostHi. Gemma-4-31B costs
\GemmaCost per 1{,}000 essays, \CostRatioBestGemma$\times$ less than \BestModel.

On the first 50 essays, the best of the \NProp proprietary models, \ClaudeBest, attains QWK
\ClaudeBestQWK [\ClaudeBestLo, \ClaudeBestHi] and Claude Opus~5.5 attains \OpusQWK, close to
human--human agreement, but their intervals overlap those of the best open-weight models on
the same essays (e.g., \SubOpenBest at \SubOpenBestQWK; Appendix~\ref{app:extra}).

\subsection{Error Structure}

\begin{figure*}[t]
  \centering
  \includegraphics{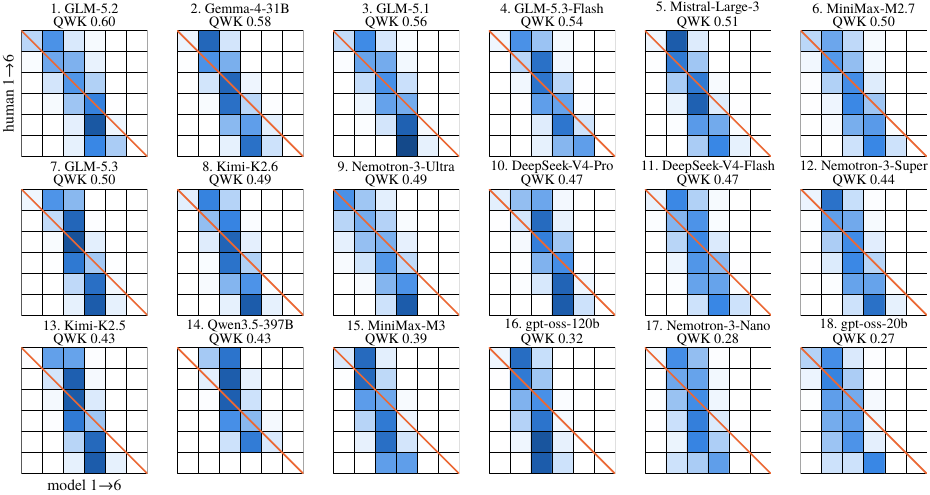}
  \caption{\textbf{Confusion matrices of all open-weight models.} Rows are human scores
  (1--6, top to bottom) and columns are model scores (1--6, left to right), normalized by
  row. Panels are numbered by QWK rank in reading order. The orange diagonal marks perfect agreement;
  mass shifted toward the central columns indicates score compression.}
  \label{fig:confusion}
\end{figure*}

Every model compresses the score scale (\RQ{2}). Of all predictions, \ModelMid are 2, 3, or
4, compared with \HumanMid of human scores, and the spread of the median model's predictions
is \SDRatio$\times$ that of human scores. Essays scored 6 by human raters receive a mean
model score of \PooledSix, and essays scored 1 receive \PooledOne
(Figure~\ref{fig:results}, right, and Figure~\ref{fig:confusion}). For the median model, only \RecHigh of
the \NHigh essays that human raters scored 5 or 6, and \RecLow of the \NLow essays scored 1,
receive a matching extreme score. Mean bias conceals this pattern: Kimi-K2.6 has a bias of
\KimiBias yet compresses scores like the other models. \citet{floden2025grading} report a similar pattern for ChatGPT on university exams.

Unlike compression, the direction of bias differs between models, from \BiasMin
(\BiasMinModel) to \BiasMax (\BiasMaxModel), and even within one family (\DSFlashBias for
DeepSeek-V4-Flash, \DSProBias for V4-Pro); \NBiasNeg of \NOpen models score below human
raters on average. Reports that LLMs score more harshly
\citep{kundu2024essay} or more leniently \citep{sessler2025grade} than human raters thus
describe particular models rather than LLM scoring in general.

Large discrepancies are uncommon for individual models but concentrated on the same essays.
Between \SevMin and \SevMax of essays differ from the human score by two or more points (median
\SevMed; \SevBest for \BestModel), and \NCons essays receive such a discrepancy from at least
two-thirds of the models. Of these, \NConsUnder are under-scored and \NConsOver over-scored;
\ConsHumanHigh have a human score of 5 or 6, the under-scored essays are long (median
\ConsUnderWordsMed words, compared with \SampleWordsMed in the sample), and the over-scored
essays are short (median \ConsOverWordsMed words).

These essays point to the second shared pattern, length insensitivity. Human scores correlate
strongly with essay length ($\rho=\RhoLenHuman$), whereas model scores correlate much less
(median $\rho=\RhoLenPredMed$, range \RhoLenPredMin--\RhoLenPredMax). Signed error therefore
decreases with length for all \NLenErrNeg models (median $\rho=\RhoLenErrMed$): in the
shortest quintile ($\leq$\WordsQOneHi words), the median model over-scores by \ErrQOne points,
and in the longest ($\geq$\WordsQFiveLo words), it under-scores by \ErrQFiveAbs points
(Appendix Figure~\ref{fig:length}). The mean human score rises from \HumanQOne to \HumanQFive across
quintiles, and models follow only part of this rise. This mirrors the length dependence
reported for small local models \citep{son2026sub3b} and runs opposite to the verbosity bias
of LLM judges \citep{zheng2023judging}. Model scores nevertheless carry information beyond
length: controlling for length lowers their median correlation with human scores from
\RhoRawMed to \RhoPartialMed, but it remains positive for every model (range
\RhoPartialMin--\RhoPartialMax), and essay length still correlates with human scores at
\RhoLenGivenLLM after controlling for the median model score.

\begin{table*}[t]
  \centering
  \begin{tabular}{@{}lrrrrrrcr@{}}
\toprule
Essay prompt & $n$ & Words & Human & $\rho_\ell^{H}$ & $\rho_\ell^{M}$ & QWK $\uparrow$ & Range & Bias $\rightarrow\!0$ \\
\midrule
Car-free cities & 89 & 436 & 2.92 & 0.60 & 0.52 & \textbf{0.58} & [0.36, 0.70] & +0.12 \\
Electoral College & 110 & 370 & 3.06 & 0.72 & 0.47 & 0.53 & [0.21, 0.63] & $-$0.24 \\
Seagoing Cowboys & 80 & 296 & 2.52 & 0.58 & 0.33 & 0.49 & [0.13, 0.60] & +0.11 \\
Facial action coding & 137 & 352 & 3.04 & 0.80 & 0.45 & 0.47 & [0.20, 0.62] & $-$0.07 \\
Face on Mars & 117 & 333 & 3.04 & 0.76 & 0.44 & 0.43 & [0.17, 0.53] & \textbf{0.00} \\
Exploring Venus & 184 & 317 & 2.72 & 0.72 & 0.38 & 0.41 & [0.17, 0.55] & $-$0.01 \\
Driverless cars & 184 & 380 & 3.23 & 0.77 & 0.37 & 0.38 & [0.14, 0.54] & $-$0.25 \\
\bottomrule
\end{tabular}

  \caption{\textbf{Agreement by essay prompt.} Results for keyword-assigned essays, sorted by
  median QWK. Words: median essay length;
  Human: mean human score; $\rho_\ell^{H}$ and $\rho_\ell^{M}$: Spearman correlation of essay
  length with human scores and with model scores (median over models); QWK: median across the
  \NOpen models; Range: minimum and maximum QWK across models; Bias: mean difference between model and
  human scores (model minus human), median over models.}
  \label{tab:prompts}
\end{table*}

Agreement also varies by essay prompt, and length explains much of this variation. Median QWK
ranges from \PromptWorstQWK on \textit{\PromptWorst} to \PromptBestQWK on
\textit{\PromptBest} (Table~\ref{tab:prompts}), a spread as large as the gap between
mid-ranked and top models and consistent with the context dependence found across studies
\citep{li2025agreement}. Human scores depend on essay length to different degrees across
prompts ($\rho$ from \RhoLenHumanPromptMin to \RhoLenHumanPromptMax), whereas model scores
relate to length similarly for all prompts. The difference between the two correlations
(\GapMin--\GapMax) is strongly associated with prompt-level agreement ($\rho=\RhoGap$,
$p=\PGap$, $n=7$ prompts): models agree best with human raters where human scores depend
least on length.

\subsection{Post-Hoc Calibration and a Length Baseline}

Model scores agree with each other only slightly more than with human raters (\RQ{3}).
Pairwise QWK between models has a median of \MMMed (range \MMMin--\MMMax;
Appendix~\ref{app:extra}), compared with a median of \MedianModelHuman with human raters.
Agreement is higher within families (median \MMWithin) than across them (\MMAcross), and a
model's mean agreement with the other models is strongly associated with its agreement with
human raters ($\rho=\RhoConv$, $p<\PConv$): stronger models converge on a shared judgment,
whereas weaker models disagree with all others. Because all models compress scores,
ensembling does not help: the median score of the three best models reaches \EnsThree, of the
five best \EnsFive, and of all \NHighParse fully parsed models \EnsAll, none higher than
\BestModel alone.

\begin{figure}[t]
  \centering
  \includegraphics{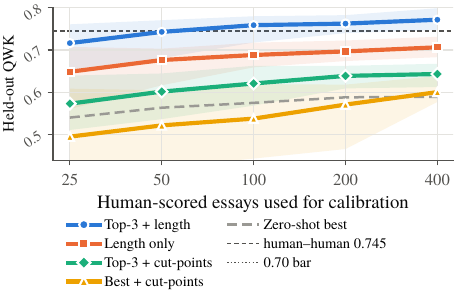}
  \caption{\textbf{Agreement after post-hoc calibration.} Each line is a scoring method whose
  mapping to the 1--6 scale is learned from the number of human-scored essays on the x-axis;
  the y-axis is QWK on the remaining essays (mean and 95\% range over \CalReps random
  splits). The dashed gray line is the best model without calibration.}
  \label{fig:calibration}
\end{figure}

Essay length alone outperforms every LLM. Calibrated on 100 human-scored essays, the
essay-length baseline reaches a held-out QWK of \CalLenHundred (Figure~\ref{fig:calibration};
Appendix~\ref{app:calib}), above every zero-shot model and above calibrated model scores
(\CalSingleHundred for the best model and \CalTopThreeHundred for the mean of the three best
models). Calibrating a single model with cut-points helps little, because integer scores leave few
thresholds to adjust; even with 400 essays it reaches only \CalSingleFourHundred.

LLM scores and essay length are complementary. Adding log essay length to the mean of the
three best models yields \CalCombTwentyFive with 25 human-scored essays, \CalCombHundred with
100, and \CalCombFourHundred [\CalCombFourHundredLo, \CalCombFourHundredHi] with 400, above
the 0.70 threshold and comparable to human--human agreement. LLM scores therefore contain
quality information that length lacks but are poorly anchored to the human scale; a
regression with three coefficients and five cut-points, fit on a few dozen local essays,
provides this anchoring.

\begin{table}[t]
  \centering
  \setlength{\tabcolsep}{1.4pt}
  \begin{tabular}{@{}lrrrrr@{}}
\toprule
Scorer & QWK $\uparrow$ & R$_{5\text{--}6}$ $\uparrow$ & R$_{1}$ $\uparrow$ & Disc.\ $\downarrow$ & \shortstack[r]{Mid\\$\rightarrow\!$85\%} \\
\midrule
ZS & 0.575 & 9\% & 21\% & 6.6\% & 97\% \\
Length & 0.688 & 58\% & 35\% & 6.1\% & 81\% \\
Best+cuts & 0.538 & 53\% & \textbf{54\%} & 19.6\% & 68\% \\
Top-3+cuts & 0.621 & 49\% & 52\% & 10.2\% & 80\% \\
Top-3+length & \textbf{0.758} & \textbf{71\%} & 45\% & \textbf{3.2\%} & \textbf{82\%} \\
\midrule
Human & -- & -- & -- & -- & 85\% \\
\bottomrule
\end{tabular}

  \caption{\textbf{Calibration with 100 human-scored essays.} ZS: best model, no calibration;
  Top-3: mean of the three best models; cuts: cut-points. R$_{5\text{--}6}$, R$_1$: share of
  extreme essays given the correct score; Mid: share of predictions at 2--4.}
  \label{tab:calib100}
\end{table}

Calibration changes where predictions fall, not only how well they correlate with human
scores (Table~\ref{tab:calib100}). With 100 human-scored essays, the combined method assigns a
5 or 6 to \CalCombRecHigh of essays that human raters scored that high, compared with
\CalRawRecHigh for the best zero-shot model, and a 1 to \CalCombRecLow of essays scored 1
(compared with \CalRawRecLow). Its share of predictions at scores 2--4 (\CalCombMid) approaches
the human share of \HumanMid, and the rate of large discrepancies decreases from \CalRawSev to
\CalCombSev. A single model with cut-points also reaches the extreme scores, but with many
large discrepancies.

Calibration largely transfers across essay prompts, which matters because teachers cannot
always score essays for each new assignment. Fit on six prompts and applied to the seventh,
the combined method averages \LopoComb (range \LopoCombMin--\LopoCombMax) and reaches 0.70 on
\NLopoCombAbove of seven prompts, compared with \LopoRaw for the best zero-shot model and
\LopoLen for the length baseline (Appendix Table~\ref{tab:lopo}). The exception is \textit{Car-free
cities}, whose essays are the longest in the sample: here length transfers poorly (QWK
\LopoLenMin) and lowers the combined method below the zero-shot model. Length--score relations are prompt-specific, so length terms should be fit per prompt when
human-scored essays are available; LLM scores transfer more reliably.

\section{Discussion}

Model size, release date (newer versions within a family), and ensembling leave score
compression and length insensitivity in place. Both are failures to anchor predictions to the
human score scale rather than of reading comprehension, since a small calibration set corrects
most of the gap, in line with gains from few-shot example essays
\citep{yancey2023rating, idowu2026specialists}, min-max scaling \citep{lee2024unleashing},
bias correction \citep{kucia2026holistic}, and light fine-tuning \citep{gayed2026l1bias}. The most cost-effective investment is thus a few dozen locally scored essays, not a larger model.

Compression appears in every family, size, and release, so it is unlikely to stem from one
training procedure, and is strongest at the rare extremes (\HumanHighPct of human scores are 5 or
6), where the rubric provides the fewest anchors, consistent with reports that scoring systems
struggle with high-quality essays \citep{idowu2026specialists}; example essays at each score
level \citep{yancey2023rating} or output scaling \citep{lee2024unleashing} target this directly.

The essay-length baseline is a validity check rather than a goal: ASAP~2.0 scores depend
strongly on length, and QWK rewards reproducing rater behavior, including reliance on length
\citep{kundu2024essay}. Models that weigh length less may be closer to the construct the rubric
describes, and adding length to LLM scores partly imports this rater behavior. We therefore do
not recommend length as a scoring feature for consequential decisions, but recommend reporting
a calibrated length baseline to show how much agreement exceeds what length alone explains.

In practice, a school can score all essays zero-shot with a mid-sized, self-hosted model, have
teachers score a few dozen essays per prompt, fit the combination of LLM score and length on
those essays, and route predicted extreme scores to human review. This reaches human--human
agreement on our data with a single model call per essay, unlike multi-agent designs that
multiply model calls and tokens \citep{jang2025roundtable}. For evaluation, we recommend reporting the predicted
score distribution, a calibrated length baseline, per-prompt results, parse rates, token usage
and cost under a documented serving setup \citep{manakina2026bandit, gui2026costaware}, and confidence
intervals, which at 50 essays have a median half-width of \SubCIHalf QWK.

\section{Limitations}

We use one dataset (U.S.\ grades 6--10, English, one holistic rubric), one prompt, and one
greedy run per model, so prompt effects cannot be separated from model effects. Open-weight
models were served by a shared hosted service, so quantization and hardware are outside our
control and model versions may change; several parameter counts are unverified
(Table~\ref{tab:main}). Token counts (first \NTokEssays essays)
depend on tokenizer and reasoning settings and are priced at list prices that change over time. Proprietary models were run on only 50 essays. Essay prompt labels are heuristic and miss about 10\% of essays.
Calibration uses repeated splits of the same 1{,}000 essays rather than an independent test set,
the human--human reference of 0.745 is reported for the full dataset, and training-data
contamination cannot be ruled out.

\section{Conclusion}
\label{sec:conclusion}

Across \NModels LLMs on ASAP~2.0, no open-weight model reaches the operational threshold of
QWK 0.70, and neither model size nor token cost predicts agreement. All models share score compression and length insensitivity, which
explains where they fail and why a few dozen human-scored essays suffice to bring LLM scores to
human--human agreement. Practitioners should therefore favor mid-sized, self-hostable models and local calibration over
scale. Future work will study with instructors how AI-assisted grading can benefit them.

\section*{Ethics and AI Usage Statement}

We use a public research dataset and do not redistribute essay text. Score compression misjudges
the strongest and weakest writers, and length-based calibration can reward verbosity, so LLM
scores should not inform consequential decisions without human oversight. An AI assistant
(Claude Code) helped write the analysis code and revise the text; the authors verified all code,
results, and claims.

\bibliography{references}

\begin{thebibliography}{38}
\providecommand{\natexlab}[1]{#1}

\bibitem[{Chiang and Lee(2023)}]{chiang2023large}
Cheng-Han Chiang and Hung-yi Lee. 2023.
\newblock \href {https://doi.org/10.18653/v1/2023.acl-long.870} {{Can Large
  Language Models Be an Alternative to Human Evaluations?}}
\newblock In \emph{Proceedings of the 61st Annual Meeting of the {A}ssociation
  for {C}omputational {L}inguistics (Volume 1: Long Papers)}, pages
  15607--15631, Toronto, Canada. Association for Computational Linguistics.

\bibitem[{Chu et~al.(2025)Chu, Li, Yang, Copur-Gencturk, and
  Tang}]{chu2025gradehitl}
Yucheng Chu, Hang Li, Kaiqi Yang, Yasemin Copur-Gencturk, and Jiliang Tang.
  2025.
\newblock \href {https://doi.org/10.1109/TALE66047.2025.11346771} {{{LLM}-Based
  Automated Grading with Human-in-the-Loop}}.
\newblock In \emph{2025 IEEE International Conference on Teaching, Assessment,
  and Learning for Engineering (TALE)}, pages 1--8.

\bibitem[{Crossley et~al.(2025)Crossley, Baffour, Burleigh, and
  King}]{crossley2025asap2}
Scott~A. Crossley, Perpetual Baffour, L.~Burleigh, and Jules King. 2025.
\newblock \href {https://doi.org/10.1016/j.asw.2025.100954} {{A large-scale
  corpus for assessing source-based writing quality: {ASAP} 2.0}}.
\newblock \emph{Assessing Writing}, 65:100954.

\bibitem[{Crossley et~al.(2022)Crossley, Baffour, Tian, Picou, Benner, and
  Boser}]{crossley2022persuade}
Scott~A. Crossley, Perpetual Baffour, Yu~Tian, Aigner Picou, Meg Benner, and
  Ulrich Boser. 2022.
\newblock \href {https://doi.org/10.1016/j.asw.2022.100667} {{The persuasive
  essays for rating, selecting, and understanding argumentative and discourse
  elements ({PERSUADE}) corpus 1.0}}.
\newblock \emph{Assessing Writing}, 54:100667.

\bibitem[{{DeepSeek-AI}(2026)}]{deepseek2026v4}
{DeepSeek-AI}. 2026.
\newblock {DeepSeek-V4: Towards Highly Efficient Million-Token Context
  Intelligence}.
\newblock \emph{arXiv preprint arXiv:2606.19348}.

\bibitem[{Flod{\'e}n(2025)}]{floden2025grading}
Jonas Flod{\'e}n. 2025.
\newblock \href {https://doi.org/10.1002/berj.4069} {{Grading exams using large
  language models: A comparison between human and {AI} grading of exams in
  higher education using {ChatGPT}}}.
\newblock \emph{British Educational Research Journal}, 51(1):201--224.

\bibitem[{Frederick~Eneye et~al.(2025)Frederick~Eneye, Ijezue, Imam~Amjad,
  Amjad, Butt, and Casta{\~n}eda-Garza}]{autograding2025survey}
Tania Amanda~Nkoyo Frederick~Eneye, Chukwuebuka~Fortunate Ijezue, Ahmad
  Imam~Amjad, Maaz Amjad, Sabur Butt, and Gerardo Casta{\~n}eda-Garza. 2025.
\newblock \href {https://doi.org/10.18653/v1/2025.bea-1.35} {{Advances in
  Auto-Grading with Large Language Models: A Cross-Disciplinary Survey}}.
\newblock In \emph{Proceedings of the 20th Workshop on Innovative Use of {NLP}
  for Building Educational Applications ({BEA} 2025)}, pages 477--498, Vienna,
  Austria. Association for Computational Linguistics.

\bibitem[{Gayed(2026)}]{gayed2026l1bias}
John~Maurice Gayed. 2026.
\newblock {Investigating first-language bias in LLM-based automated essay
  scoring: A cross-prompt evaluation of an open-weight AI-model on TOEFL
  essays}.
\newblock \emph{arXiv preprint arXiv:2607.14605}.

\bibitem[{{Gemma Team}(2026)}]{gemma2026gemma4}
{Gemma Team}. 2026.
\newblock {Gemma 4 Technical Report}.
\newblock \emph{arXiv preprint arXiv:2607.02770}.

\bibitem[{{GLM-5 Team}(2026)}]{glm2026glm5}
{GLM-5 Team}. 2026.
\newblock {GLM-5: from Vibe Coding to Agentic Engineering}.
\newblock \emph{arXiv preprint arXiv:2602.15763}.

\bibitem[{Gui(2026)}]{gui2026costaware}
Yi~Gui. 2026.
\newblock {From Efficiency to Sufficiency: The Cost-Aware Measurement Design
  Framework for Automated Essay Scoring}.
\newblock \emph{arXiv preprint arXiv:2610.19287}.

\bibitem[{Hua(2026)}]{hua2026summarization}
Haowei Hua. 2026.
\newblock {Cost-efficient generative AI summarization for scalable automated
  essay scoring in educational assessment}.
\newblock \emph{arXiv preprint arXiv:2607.15829}.

\bibitem[{Idowu and Almasoud(2026)}]{idowu2026specialists}
Jamiu~Adekunle Idowu and Ahmed Almasoud. 2026.
\newblock {Specialists or Generalists? Multi-Agent and Single-Agent LLMs for
  Essay Grading}.
\newblock \emph{arXiv preprint arXiv:2601.22386}.

\bibitem[{Jang et~al.(2025)Jang, Moon, Jung, Kim, and Lee}]{jang2025roundtable}
Jinhee Jang, Ayoung Moon, Minkyoung Jung, YoungBin Kim, and Seung~Jin Lee.
  2025.
\newblock {LLM Agents at the Roundtable: A Multi-Perspective and Dialectical
  Reasoning Framework for Essay Scoring}.
\newblock \emph{arXiv preprint arXiv:2509.14834}.

\bibitem[{{Kimi Team}(2026)}]{kimi2025k2}
{Kimi Team}. 2026.
\newblock {Kimi K2: Open Agentic Intelligence}.
\newblock \emph{arXiv preprint arXiv:2507.20534}.

\bibitem[{Kucia et~al.(2026)Kucia, Chakraborty, and
  Wróblewska}]{kucia2026holistic}
Filip~J. Kucia, Anirban Chakraborty, and Anna Wróblewska. 2026.
\newblock \href {https://doi.org/10.1007/978-3-032-29918-5_38} {{LLM Essay
  Scoring Under Holistic and Analytic Rubrics: Prompt Effects and Bias}}.
\newblock In \emph{Computational Science -- ICCS 2026 Workshops}, volume 16789
  of \emph{Lecture Notes in Computer Science}, pages 531--546. Springer, Cham.

\bibitem[{Kundu and Barbosa(2024)}]{kundu2024essay}
Anindita Kundu and Denilson Barbosa. 2024.
\newblock {Are Large Language Models Good Essay Graders?}
\newblock \emph{arXiv preprint arXiv:2409.13120}.

\bibitem[{Lee et~al.(2024)Lee, Cai, Meng, Wang, and Wu}]{lee2024unleashing}
Sanwoo Lee, Yida Cai, Desong Meng, Ziyang Wang, and Yunfang Wu. 2024.
\newblock \href {https://doi.org/10.18653/v1/2024.findings-emnlp.10}
  {{Unleashing Large Language Models' Proficiency in Zero-shot Essay Scoring}}.
\newblock In \emph{Findings of the {A}ssociation for {C}omputational
  {L}inguistics: {EMNLP} 2024}, pages 181--198, Miami, Florida, USA.
  Association for Computational Linguistics.

\bibitem[{Li et~al.(2026)Li, Chen, Fan, Young-Johnson, Lim, and
  Feng}]{li2025agreement}
Hongli Li, Che~Han Chen, Kevin Fan, Chiho Young-Johnson, Soyoung Lim, and Yali
  Feng. 2026.
\newblock {Agreement Between Large Language Models and Human Raters in Essay
  Scoring: A Research Synthesis}.
\newblock \emph{arXiv preprint arXiv:2512.14561}.

\bibitem[{Manakina and Bogdanov(2026)}]{manakina2026bandit}
Olga Manakina and Igor Bogdanov. 2026.
\newblock {Learning to Grade Efficiently: A Bandit-Driven Prompt-Selection
  Framework for Low-Cost LLM Essay Scoring}.
\newblock \emph{arXiv preprint arXiv:2608.23814}.

\bibitem[{{MiniMax}(2026)}]{minimax2026m2}
{MiniMax}. 2026.
\newblock {The MiniMax-M2 Series: Mini Activations Unleashing Max Real-World
  Intelligence}.
\newblock \emph{arXiv preprint arXiv:2605.26494}.

\bibitem[{{NVIDIA}(2025)}]{nvidia2025nemotron3}
{NVIDIA}. 2025.
\newblock {NVIDIA Nemotron 3: Efficient and Open Intelligence}.
\newblock \emph{arXiv preprint arXiv:2512.20856}.

\bibitem[{{OpenAI}(2025)}]{openai2025gptoss}
{OpenAI}. 2025.
\newblock {gpt-oss-120b \& gpt-oss-20b Model Card}.
\newblock \emph{arXiv preprint arXiv:2508.10925}.

\bibitem[{Ormerod and Kehat(2025)}]{ormerod2025long}
Christopher Ormerod and Gitit Kehat. 2025.
\newblock {Long Context Automated Essay Scoring with Language Models}.
\newblock \emph{arXiv preprint arXiv:2509.10417}.

\bibitem[{Seßler et~al.(2025)Seßler, Fürstenberg, Bühler, and
  Kasneci}]{sessler2025grade}
Kathrin Seßler, Maurice Fürstenberg, Babette Bühler, and Enkelejda Kasneci.
  2025.
\newblock \href {https://doi.org/10.1145/3706468.3706527} {{Can AI grade your
  essays? A comparative analysis of large language models and teacher ratings
  in multidimensional essay scoring}}.
\newblock In \emph{Proceedings of the 15th International Learning Analytics and
  Knowledge Conference (LAK '25)}, pages 462--472.

\bibitem[{Shibata and Miyamura(2025)}]{shibata2025lces}
Takumi Shibata and Yuichi Miyamura. 2025.
\newblock \href {https://doi.org/10.18653/v1/2025.emnlp-main.1523} {{{LCES}:
  Zero-shot Automated Essay Scoring via Pairwise Comparisons Using Large
  Language Models}}.
\newblock In \emph{Proceedings of the 2025 Conference on Empirical Methods in
  Natural Language Processing}, pages 29988--30001, Suzhou, China. Association
  for Computational Linguistics.

\bibitem[{Son et~al.(2026)Son, Minh, Loi, Vu, and Anh}]{son2026sub3b}
Nguyen~Dung Son, Dang~Quang Minh, Nguyen~Huu Loi, Truong~Viet Vu, and
  Nguyen~Thai Anh. 2026.
\newblock {How Far Can Sub-3B Open Language Models Go in Zero-Shot Essay
  Scoring on an 8 GB Consumer GPU?}
\newblock \emph{arXiv preprint arXiv:2609.20250}.

\bibitem[{Stahl et~al.(2024)Stahl, Biermann, Nehring, and
  Wachsmuth}]{stahl2024exploring}
Maja Stahl, Leon Biermann, Andreas Nehring, and Henning Wachsmuth. 2024.
\newblock \href {https://aclanthology.org/2024.bea-1.23/} {{Exploring {LLM}
  Prompting Strategies for Joint Essay Scoring and Feedback Generation}}.
\newblock In \emph{Proceedings of the 19th Workshop on Innovative Use of {NLP}
  for Building Educational Applications ({BEA} 2024)}, pages 283--298, Mexico
  City, Mexico. Association for Computational Linguistics.

\bibitem[{Su et~al.(2025)Su, Yan, Fu, Han, Ye, Liu, Huo, Zhou, and
  Hu}]{su2025essayjudge}
Jiamin Su, Yibo Yan, Fangteng Fu, Zhang Han, Jingheng Ye, Xiang Liu, Jiahao
  Huo, Huiyu Zhou, and Xuming Hu. 2025.
\newblock \href {https://doi.org/10.18653/v1/2025.findings-acl.329}
  {{{E}ssay{J}udge: A Multi-Granular Benchmark for Assessing Automated Essay
  Scoring Capabilities of Multimodal Large Language Models}}.
\newblock In \emph{Findings of the {A}ssociation for {C}omputational
  {L}inguistics: {ACL} 2025}, pages 6363--6389, Vienna, Austria. Association
  for Computational Linguistics.

\bibitem[{Taghipour and Ng(2016)}]{taghipour2016neural}
Kaveh Taghipour and Hwee~Tou Ng. 2016.
\newblock \href {https://doi.org/10.18653/v1/D16-1193} {{A Neural Approach to
  Automated Essay Scoring}}.
\newblock In \emph{Proceedings of the 2016 Conference on Empirical Methods in
  Natural Language Processing}, pages 1882--1891, Austin, Texas. Association
  for Computational Linguistics.

\bibitem[{Tam et~al.(2024)Tam, Wu, Tsai, Lin, Lee, and Chen}]{tam2024speak}
Zhi~Rui Tam, Cheng-Kuang Wu, Yi-Lin Tsai, Chieh-Yen Lin, Hung-yi Lee, and
  Yun-Nung Chen. 2024.
\newblock \href {https://doi.org/10.18653/v1/2024.emnlp-industry.91} {{Let Me
  Speak Freely? A Study On The Impact Of Format Restrictions On Large Language
  Model Performance.}}
\newblock In \emph{Proceedings of the 2024 Conference on Empirical Methods in
  Natural Language Processing: Industry Track}, pages 1218--1236, Miami,
  Florida, US. Association for Computational Linguistics.

\bibitem[{Wang et~al.(2026)Wang, Ding, Wu, Sun, Liu, and
  Zhai}]{wang2026autoscore}
Yun Wang, Zhaojun Ding, Xuansheng Wu, Siyue Sun, Ninghao Liu, and Xiaoming
  Zhai. 2026.
\newblock \href {https://doi.org/10.1609/aaai.v40i48.42123} {{AutoSCORE:
  Enhancing Automated Scoring with Multi-Agent Large Language Models via
  Structured Component Recognition}}.
\newblock In \emph{Proceedings of the AAAI Conference on Artificial
  Intelligence}, volume~40, pages 40898--40906.

\bibitem[{Willard and Louf(2023)}]{willard2023efficient}
Brandon~T. Willard and Rémi Louf. 2023.
\newblock {Efficient Guided Generation for Large Language Models}.
\newblock \emph{arXiv preprint arXiv:2307.09702}.

\bibitem[{Williamson et~al.(2012)Williamson, Xi, and
  Breyer}]{williamson2012framework}
David~M. Williamson, Xiaoming Xi, and F.~Jay Breyer. 2012.
\newblock \href {https://doi.org/10.1111/j.1745-3992.2011.00223.x} {{A
  Framework for Evaluation and Use of Automated Scoring}}.
\newblock \emph{Educational Measurement: Issues and Practice}, 31(1):2--13.

\bibitem[{Xiao et~al.(2025)Xiao, Ma, Song, Xu, Zhang, Wang, and
  Fu}]{xiao2025human}
Changrong Xiao, Wenxing Ma, Qingping Song, Sean~Xin Xu, Kunpeng Zhang, Yufang
  Wang, and Qi~Fu. 2025.
\newblock \href {https://doi.org/10.1145/3706468.3706507} {{Human-AI
  Collaborative Essay Scoring: A Dual-Process Framework with LLMs}}.
\newblock In \emph{Proceedings of the 15th International Learning Analytics and
  Knowledge Conference (LAK '25)}, pages 293--305.

\bibitem[{Yancey et~al.(2023)Yancey, Laflair, Verardi, and
  Burstein}]{yancey2023rating}
Kevin~P. Yancey, Geoffrey Laflair, Anthony Verardi, and Jill Burstein. 2023.
\newblock \href {https://doi.org/10.18653/v1/2023.bea-1.49} {{Rating Short {L}2
  Essays on the {CEFR} Scale with {GPT}-4}}.
\newblock In \emph{Proceedings of the 18th Workshop on Innovative Use of {NLP}
  for Building Educational Applications ({BEA} 2023)}, pages 576--584, Toronto,
  Canada. Association for Computational Linguistics.

\bibitem[{Yoshida(2025)}]{yoshida2025detailed}
Lui Yoshida. 2025.
\newblock \href {https://doi.org/10.1007/978-3-031-98465-5_8} {{Do We Need a
  Detailed Rubric for Automated Essay Scoring Using Large Language Models?}}
\newblock In \emph{Artificial Intelligence in Education (AIED 2025)}, volume
  15882 of \emph{Lecture Notes in Computer Science}, pages 60--67. Springer,
  Cham.

\bibitem[{Zheng et~al.(2023)Zheng, Chiang, Sheng, Zhuang, Wu, Zhuang, Lin, Li,
  Li, Xing, Zhang, Gonzalez, and Stoica}]{zheng2023judging}
Lianmin Zheng, Wei-Lin Chiang, Ying Sheng, Siyuan Zhuang, Zhanghao Wu, Yonghao
  Zhuang, Zi~Lin, Zhuohan Li, Dacheng Li, Eric~P. Xing, Hao Zhang, Joseph~E.
  Gonzalez, and Ion Stoica. 2023.
\newblock \href {https://doi.org/10.52202/075280-2020} {{Judging LLM-as-a-Judge
  with MT-Bench and Chatbot Arena}}.
\newblock In \emph{Advances in Neural Information Processing Systems 36
  (Datasets and Benchmarks Track)}, pages 46595--46623.

\end{thebibliography}

\clearpage
\appendix

\section{Prompt and Output Schema}
\label{app:prompt}

The system role is: \textit{You are an expert essay rater. Score the student essay
holistically on an integer scale from 1 (minimum) to 6 (maximum) using ONLY the rubric
provided. Return your answer as JSON with an integer `score' and a one-sentence
`rationale'.} The user message is: \texttt{\# Holistic Scoring Rubric} $\langle$official
ASAP~2.0 holistic rubric, six score descriptors$\rangle$ \texttt{---} \texttt{\# Essay to
score} $\langle$essay$\rangle$ \texttt{---} \textit{Assign one integer holistic score from
1 to 6.} The output schema is \texttt{\{"score": integer (min 1, max 6), "rationale":
string\}}, with \texttt{score} required. The proprietary API's structured-output mode does not
support numeric bounds, so out-of-range scores are rejected after decoding.

\section{Models and Token Costs}
\label{app:models}

Table~\ref{tab:models} lists every evaluated model with its serving tag, architecture, and
parameter counts, together with the source of each count. Only sizes marked \emph{report} are
stated in a technical report for that exact model; sizes marked \emph{family} come from the
family's report, and \emph{unverified} sizes are not confirmed by a cited source, which is why
Table~\ref{tab:main} flags them.

Table~\ref{tab:prices} gives the inputs of the cost estimate in Section~4.1. For each model,
the cost of scoring 1{,}000 essays is $1{,}000\,(\bar t_\mathrm{in}\,p_\mathrm{in} + \bar
t_\mathrm{out}\,p_\mathrm{out})$, where $\bar t$ is the mean number of input or output tokens
per essay over the first \NTokEssays essays and $p$ is the list price per token. Input tokens
differ little between models, because every request contains the same rubric and essay;
differences in cost therefore come from output tokens, which include reasoning, and from
price. The \NTableOmit models below the line were retired from our hosted service before
token usage was measured.

\begin{table*}[tp]
  \centering
  \setlength{\tabcolsep}{6pt}
  \begin{tabular}{@{}lrrrrr@{}}
\toprule
& \multicolumn{2}{c}{List price (\$/M tok.)} & \multicolumn{2}{c}{Tokens per essay} & \\
\cmidrule(lr){2-3}\cmidrule(lr){4-5}
Model & Input & Output & Input & Output & \$/1k essays \\
\midrule
GLM-5.2 & 0.620 & 2.420 & 1{,}305 & 423 & 1.83 \\
Gemma-4-31B & 0.200 & 0.400 & 1{,}315 & 62 & 0.29 \\
GLM-5.3-Flash & 0.120 & 0.380 & 1{,}305 & 830 & 0.47 \\
GLM-5.3 & 0.070 & 1.400 & 1{,}305 & 1{,}181 & 1.74 \\
Kimi-K2.6 & 0.650 & 3.410 & 1{,}304 & 1{,}862 & 7.20 \\
Nemotron-3-Ultra & 0.500 & 2.200 & 1{,}334 & 455 & 1.67 \\
DeepSeek-V4-Pro & 0.888 & 1.508 & 1{,}297 & 814 & 2.38 \\
Nemotron-3-Super & 0.075 & 0.400 & 1{,}334 & 1{,}152 & 0.56 \\
MiniMax-M3 & 0.300 & 1.200 & 1{,}468 & 505 & 1.05 \\
gpt-oss-120b & 0.050 & 0.250 & 1{,}362 & 329 & 0.15 \\
gpt-oss-20b & 0.030 & 0.130 & 1{,}362 & 580 & 0.12 \\
\midrule
GLM-5.1 & 1.400 & 4.400 & -- & -- & -- \\
Mistral-Large-3 & 0.500 & 1.500 & -- & -- & -- \\
MiniMax-M2.7 & 0.600 & 1.800 & -- & -- & -- \\
DeepSeek-V4-Flash & 0.090 & 0.180 & -- & -- & -- \\
Kimi-K2.5 & 0.500 & 2.500 & -- & -- & -- \\
Qwen3.5-397B & 0.500 & 3.600 & -- & -- & -- \\
Nemotron-3-Nano & 0.060 & 0.240 & -- & -- & -- \\
\bottomrule
\end{tabular}

  \caption{\textbf{Token prices and usage.} List prices per million tokens (OpenRouter,
  retrieved 2026-11-25) and mean tokens per essay over the first \NTokEssays essays; output
  tokens include reasoning. --\,: not measured.}
  \label{tab:prices}
\end{table*}

\begin{table*}[tp]
  \centering
  \begin{tabular}{@{}llllrr@{}}
\toprule
Model & Serving tag & Arch. & Size source & Total (B) & Active (B) \\
\midrule
gpt-oss-20b & \texttt{gpt-oss:20b-cloud} & MoE & report & 20.9 & 3.6 \\
gpt-oss-120b & \texttt{gpt-oss:120b-cloud} & MoE & report & 116.8 & 5.1 \\
Gemma-4-31B & \texttt{gemma4:31b} & dense & report & 31 & 31 \\
Nemotron-3-Nano & \texttt{nemotron-3-nano:30b-cloud} & MoE & report & 30 & 3 \\
Nemotron-3-Super & \texttt{nemotron-3-super:cloud} & MoE & unverified & 120 &  \\
Nemotron-3-Ultra & \texttt{nemotron-3-ultra:cloud} & MoE & unverified & 550 &  \\
DeepSeek-V4-Flash & \texttt{deepseek-v4-flash:cloud} & MoE & report & 284 & 13 \\
DeepSeek-V4-Pro & \texttt{deepseek-v4-pro:cloud} & MoE & report & 1600 & 49 \\
MiniMax-M2.7 & \texttt{minimax-m2.7:cloud} & MoE & report & 229.9 & 9.8 \\
MiniMax-M3 & \texttt{minimax-m3:cloud} & -- & unverified & 427 &  \\
Qwen3.5-397B & \texttt{qwen3.5:397b-cloud} & -- & tag & 397 &  \\
Mistral-Large-3 & \texttt{mistral-large-3:675b-cloud} & -- & tag & 675 &  \\
GLM-5.1 & \texttt{glm-5.1:cloud} & MoE & family & 744 & 40 \\
GLM-5.2 & \texttt{glm-5.2:cloud} & MoE & family & 744 & 40 \\
GLM-5.3 & \texttt{glm-5.3:cloud} & MoE & unverified & 753 & 40 \\
GLM-5.3-Flash & \texttt{glm-5.3-flash:cloud} & MoE & unverified & 321 & 18 \\
Kimi-K2.5 & \texttt{kimi-k2.5:cloud} & MoE & family & 1040 & 32 \\
Kimi-K2.6 & \texttt{kimi-k2.6:cloud} & MoE & family & 1040 & 32 \\
Claude Haiku 4.5 & \texttt{claude-haiku-4-5} & -- & unknown &  &  \\
Claude Sonnet 5.5 & \texttt{claude-sonnet-5-5} & -- & unknown &  &  \\
Claude Fable 5.1 & \texttt{claude-fable-5-1} & -- & unknown &  &  \\
Claude Opus 5.5 & \texttt{claude-opus-5-5} & -- & unknown &  &  \\
\bottomrule
\end{tabular}

  \caption{\textbf{Evaluated models.} Citations are given in Section~3. Size source: \emph{report}, stated in the cited
  technical report for this exact model; \emph{family}, taken from the family's report (the
  point release is not described); \emph{tag}, from the serving tag name; \emph{unverified},
  not confirmed by a cited source.}
  \label{tab:models}
\end{table*}

\section{Calibration Results}
\label{app:calib}

Table~\ref{tab:calib} extends Table~\ref{tab:calib100} to every calibration set size shown in
Figure~\ref{fig:calibration}. Combining the three best models with essay length (Top-3+length)
is the best scorer at every size, from \CalCombTwentyFive with 25 human-scored essays to
\CalCombFourHundred with 400, and it stays above length alone (\CalLenTwentyFive to
\CalLenFourHundred). Length alone, in turn, outperforms both calibrations that use LLM scores
only. Calibrating the single best model with cut-points (Best+cuts) remains below the
uncalibrated model (ZS) until 400 essays, where it reaches \CalSingleFourHundred, because a
model's integer scores leave few positions for the cut-points.

Table~\ref{tab:lopo} gives the leave-one-prompt-out results summarized in Section~4.3: each
scorer is fit on the essays of six prompts and evaluated on the seventh, so no essay from the
held-out prompt is used for fitting. The combined scorer (Comb.) is best or tied for best on
every held-out prompt except \textit{Car-free cities}, where the length--score relation of the
other prompts transfers poorly.

\begin{table}[tbp]
  \centering
  \setlength{\tabcolsep}{2.2pt}
  \begin{tabular}{@{}lrrrrr@{}}
\toprule
Scorer (QWK $\uparrow$) & 25 & 50 & 100 & 200 & 400 \\
\midrule
ZS & 0.540 & 0.564 & 0.575 & 0.588 & 0.590 \\
Length & 0.648 & 0.676 & 0.688 & 0.697 & 0.706 \\
Best+cuts & 0.496 & 0.522 & 0.538 & 0.571 & 0.601 \\
Top-3+cuts & 0.573 & 0.602 & 0.621 & 0.639 & 0.643 \\
Top-3+length & \textbf{0.716} & \textbf{0.742} & \textbf{0.758} & \textbf{0.762} & \textbf{0.771} \\
\bottomrule
\end{tabular}

  \caption{\textbf{Held-out QWK by calibration set size.} Columns give the number of
  human-scored essays used for calibration; means over \CalReps random splits
  (Figure~\ref{fig:calibration}). Labels as in Table~\ref{tab:calib100}.}
  \label{tab:calib}
\end{table}

\begin{table}[tbp]
  \centering
  \setlength{\tabcolsep}{3pt}
  \begin{tabular}{@{}lrrrr@{}}
\toprule
Held-out prompt & \multicolumn{4}{c}{QWK $\uparrow$} \\
\cmidrule(l){2-5}
 & ZS & Top-3 & Length & Comb. \\
\midrule
Electoral College & 0.63 & 0.58 & 0.73 & \textbf{0.80} \\
Facial action coding & 0.62 & 0.69 & 0.77 & \textbf{0.80} \\
Driverless cars & 0.52 & 0.62 & 0.72 & \textbf{0.76} \\
Exploring Venus & 0.55 & 0.60 & 0.69 & \textbf{0.75} \\
Face on Mars & 0.51 & 0.58 & \textbf{0.75} & \textbf{0.75} \\
Seagoing Cowboys & 0.54 & 0.57 & 0.57 & \textbf{0.68} \\
Car-free cities & \textbf{0.70} & \textbf{0.70} & 0.49 & 0.63 \\
\midrule
Mean & 0.58 & 0.62 & 0.67 & \textbf{0.74} \\
\bottomrule
\end{tabular}

  \caption{\textbf{Leave-one-prompt-out QWK.} Each method is fit on six essay prompts and
  tested on the seventh. ZS: best model, no calibration; Top-3: mean of the three best models;
  Comb.: Top-3 plus essay length.}
  \label{tab:lopo}
\end{table}

\section{Essay Prompt Assignment Rules}
\label{app:prompts}

An essay is assigned to a prompt when exactly one of the following case-insensitive patterns
matches: \textit{Exploring Venus}: ``venus''; \textit{Face on Mars}: ``face on mars'' or
``face'' and ``mars''; \textit{Driverless cars}: ``driverless''; \textit{Electoral College}:
``electoral college''; \textit{Car-free cities}: ``car-free'', ``limiting car'', ``car usage'',
or ``Vauban''; \textit{Facial action coding}: ``facial action'', ``FACS'', ``emotional
expression/recognition'', or ``Mona Lisa''; \textit{Seagoing Cowboys}: ``seagoing'',
``cowboy'', or ``Luke''. The rules match \PromptCoverageTrain of the full training split.

\section{Additional Results}
\label{app:extra}

Table~\ref{tab:main_omitted} completes Table~\ref{tab:main} with the \NTableOmit open-weight
models omitted there for space. They are included in every analysis in the paper; only their
token columns are blank, because they were retired from our hosted service before the cost
measurement.

Table~\ref{tab:subset} compares all \NModels models, including the \NProp proprietary ones, on
the first 50 essays, the only essays scored by every model. \ClaudeBest and Claude Opus~5.5
rank highest, but at this sample size the 95\% intervals have a median half-width of
\SubCIHalf QWK and overlap those of the best open-weight models, so the ranking is not
reliable.

\begin{figure}[tbp]
  \centering
  \includegraphics{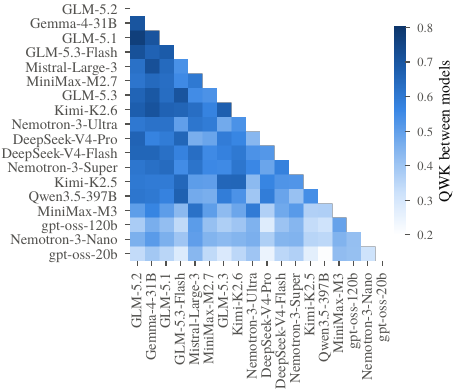}
  \caption{\textbf{Pairwise QWK between the \NOpen open-weight models.} Models are ordered by
  agreement with human raters.}
  \label{fig:agreement}
\end{figure}

\begin{figure}[tbp]
  \centering
  \includegraphics{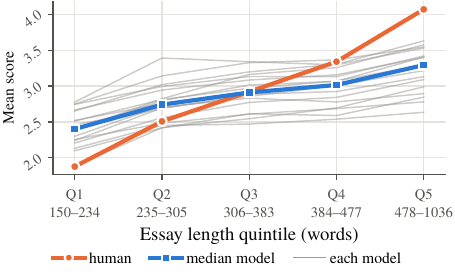}
  \caption{\textbf{Mean score by essay-length quintile.} Human scores rise steeply with essay
  length, whereas model scores rise much less, so long essays are under-scored and short
  essays over-scored.}
  \label{fig:length}
\end{figure}

\begin{figure}[tbp]
  \centering
  \includegraphics{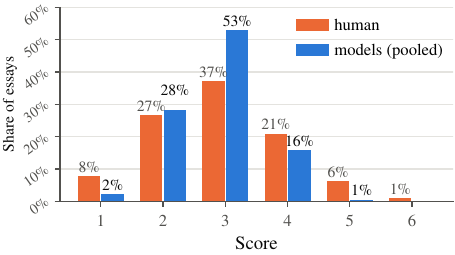}
  \caption{\textbf{Score distributions.} Human scores versus all predictions of the \NOpen
  open-weight models (pooled).}
  \label{fig:dist}
\end{figure}

\begin{figure}[tbp]
  \centering
  \includegraphics{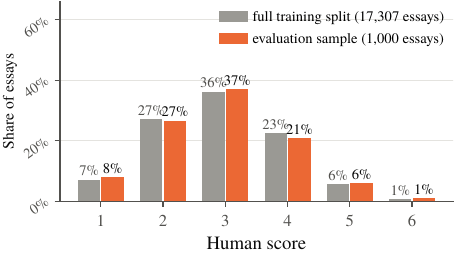}
  \caption{\textbf{Score distribution of the evaluation sample.} Share of essays at each
  human score in the full training split (\NTrain essays) and in the \NSample-essay sample
  drawn from it.}
  \label{fig:sampledist}
\end{figure}

Figure~\ref{fig:agreement} shows the agreement between pairs of open-weight models summarized
in Section~4.3. The models that agree best with human raters also agree closely with one
another, whereas the weakest models agree little with any other model.
Figure~\ref{fig:length} plots the mean human and model score in each essay-length quintile
(Section~4.2): human scores rise steeply with length and model scores much less, so models
over-score short essays and under-score long ones. Figure~\ref{fig:dist} pools all
predictions of the open-weight models: \ModelMid of them are 2, 3, or 4, compared with
\HumanMid of human scores. Finally, Figure~\ref{fig:sampledist} shows that the
\NSample-essay sample reproduces the score distribution of the full training split, within
\SampleDistMaxDiff percentage points at every score.

\begin{table*}[tp]
  \centering
  \setlength{\tabcolsep}{2.6pt}
  \begin{tabular}{@{}lrrcrrrrrr@{}}
\toprule
Model & Size (B) & QWK $\uparrow$ & 95\% CI & Exact $\uparrow$ & Disc.\ $\downarrow$ & Bias $\rightarrow\!0$ & Parsed $\uparrow$ & \shortstack[r]{Out\\tok.\ $\downarrow$} & \shortstack[r]{\$ per\\1k $\downarrow$} \\
\midrule
GLM-5.1 & 744/40$^\dagger$ & 0.565 & [0.53, 0.60] & 0.45 & 6.6\% & \textbf{$-$0.01} & \textbf{100\%} & -- & -- \\
Mistral-Large-3 & 675 & 0.511 & [0.47, 0.55] & 0.46 & 7.1\% & $-$0.25 & \textbf{100\%} & -- & -- \\
MiniMax-M2.7 & 230/9.8 & 0.502 & [0.46, 0.55] & 0.45 & 9.3\% & $-$0.14 & \textbf{100\%} & -- & -- \\
DeepSeek-V4-Flash & 284/13 & 0.467 & [0.42, 0.51] & 0.42 & 8.6\% & $-$0.04 & \textbf{100\%} & -- & -- \\
Kimi-K2.5 & 1,040/32$^\dagger$ & 0.428 & [0.39, 0.47] & 0.42 & 7.8\% & +0.14 & \textbf{100\%} & -- & -- \\
Qwen3.5-397B & 397 & 0.427 & [0.34, 0.51] & 0.39 & 10.4\% & +0.28 & 22\% & -- & -- \\
Nemotron-3-Nano & 30/3 & 0.281 & [0.23, 0.33] & 0.37 & 14.7\% & $-$0.21 & \textbf{100\%} & -- & -- \\
\bottomrule
\end{tabular}

  \caption{\textbf{Zero-shot agreement of the \NTableOmit models omitted from Table~\ref{tab:main}.}
  Columns and bold as in Table~\ref{tab:main}. Qwen3.5-397B's QWK rests on the \QwenParse of
  essays it scored. --\,: token usage not yet measured (models retired from the hosted service
  before the cost measurement; Section~3).}
  \label{tab:main_omitted}
\end{table*}

\begin{table}[tbp]
  \centering
  \setlength{\tabcolsep}{3pt}
  \begin{tabular}{@{}lrc@{}}
\toprule
Model & QWK $\uparrow$ & 95\% CI \\
\midrule
Claude Fable 5.1$^{*}$ & \textbf{0.748} & [0.59, 0.85] \\
Claude Opus 5.5$^{*}$ & 0.734 & [0.61, 0.81] \\
Claude Sonnet 5.5$^{*}$ & 0.613 & [0.41, 0.75] \\
MiniMax-M2.7 & 0.610 & [0.42, 0.76] \\
GLM-5.2 & 0.594 & [0.38, 0.75] \\
Gemma-4-31B & 0.578 & [0.37, 0.71] \\
GLM-5.1 & 0.554 & [0.34, 0.71] \\
Nemotron-3-Ultra & 0.546 & [0.35, 0.69] \\
GLM-5.3-Flash & 0.536 & [0.38, 0.66] \\
Claude Haiku 4.5$^{*}$ & 0.511 & [0.28, 0.68] \\
Qwen3.5-397B & 0.491 & [0.28, 0.64] \\
DeepSeek-V4-Flash & 0.470 & [0.24, 0.65] \\
Kimi-K2.6 & 0.458 & [0.27, 0.61] \\
Mistral-Large-3 & 0.433 & [0.24, 0.59] \\
DeepSeek-V4-Pro & 0.428 & [0.24, 0.57] \\
GLM-5.3 & 0.378 & [0.13, 0.55] \\
Nemotron-3-Super & 0.371 & [0.17, 0.56] \\
MiniMax-M3 & 0.345 & [0.15, 0.51] \\
Kimi-K2.5 & 0.333 & [0.11, 0.53] \\
gpt-oss-20b & 0.298 & [0.08, 0.49] \\
gpt-oss-120b & 0.257 & [0.07, 0.43] \\
Nemotron-3-Nano & 0.124 & [$-$0.09, 0.31] \\
\bottomrule
\end{tabular}

  \caption{\textbf{All \NModels models on the first 50 essays of the sample.} $^{*}$Proprietary model.
  Intervals are wide at this sample size, so rankings are not reliable.}
  \label{tab:subset}
\end{table}

\end{document}